\pdfoutput=1

\documentclass[11pt]{article}

\usepackage[preprint]{acl}

\usepackage{times}
\usepackage{latexsym}

\usepackage[T1]{fontenc}

\usepackage[utf8]{inputenc}

\usepackage{microtype}
\usepackage{balance}
\usepackage{tabularx, booktabs}
\usepackage{multirow}
\usepackage{makecell}
\usepackage{amssymb}
\usepackage{booktabs}
\usepackage{threeparttable}
\usepackage{subcaption}
\usepackage{tablefootnote}
\usepackage{geometry}
\usepackage{array}
\usepackage[ruled, linesnumbered]{algorithm2e}
\usepackage{xcolor}
\usepackage{textcomp}
\usepackage{graphicx}
\usepackage{amsmath}
\usepackage{acronym}

\acrodef{IR}{information retrieval}
\acrodef{MLP}{multilayer perceptron}
\acrodef{RNN}{recurrent neural network}
\acrodef{ERP}{event related potential}
\acrodef{EEG}{electroencephalogram}
\acrodef{AUC}{area under curve}
\acrodef{fMRI}{functional magnetic resonance imaging}
\acrodef{BCI}{brain–computer interface}
\acrodef{DE}{differential entropy}
\acrodef{RASM}{rational asymmetry}
\acrodef{DASM}{differential asymmetry}
\acrodef{SVM}{support vector machines}
\acrodef{DBN}{deep belief networks}
\acrodef{KNN}{k-nearest neighbors}
\acrodef{MEG}{magnetoencephalogram}
\acrodef{IDF}{inverse document frequency}
\acrodef{NDCG}{normalized discounted cumulative gain}
\acrodef{MAP}{mean average precision}
\acrodef{MSE}{mean square loss}
\acrodef{ICT}{inverse cloze test}
\acrodef{TR}{time repetition}
\acrodef{SCQ}{similarity collection-query}
\acrodef{ICTF}{inverse collection term frequency}
\acrodef{fNIRS}{Functional near Infrared Spectroscopy}

\definecolor{true_blue}{rgb}{0,0,200}
\definecolor{myPurple}{RGB}{128,0,128}
\definecolor{myLightPurple}{RGB}{200,150,200}

\AtBeginDocument{%
  \providecommand\BibTeX{{%
    \normalfont B\kern-0.5em{\scshape i\kern-0.25em b}\kern-0.8em\TeX}}}

\makeatletter
\g@addto@macro\normalsize{%
  \abovedisplayskip 2pt plus1pt 
  \belowdisplayskip 2pt plus1pt
  \abovedisplayshortskip  0pt plus1pt%
  \belowdisplayshortskip  0pt plus1pt
}
\makeatother

\newcommand{\zjtmerged}[1]{{\color{black}#1}}

\usepackage{inconsolata}
\usepackage[skip=5pt]{caption}
\usepackage[inline]{enumitem}

\PassOptionsToPackage{sort}{natbib}

\title{Query Augmentation by Decoding Semantics from Brain Signals}

\author{Ziyi Ye$^{1}$, Jingtao Zhan$^{1}$, Qingyao Ai$^{1}$, Yiqun Liu$^{1}$, \\ \textbf{Maarten de Rijke$^{2}$, Christina Lioma$^{3}$, Tuukka Ruotsalo$^{3,4}$} \\ 
{\small $^1$Quan Cheng Lab, Tsinghua University\hspace{5mm}$^2$University of Amsterdam\hspace{5mm}$^3$University of Copenhagen\hspace{5mm}$^4$LUK University} \\
{\small \texttt{\{yeziyi1998,jingtaozhan\}@gmail.com}}\hspace{3mm}
{\small \texttt{\{aiqy,yiqunliu\}@tsinghua.edu.cn}}\hspace{3mm}
{\small \texttt{M.deRijke@uva.nl}}\hspace{3mm}
{\small \texttt{\{tr,c.lioma\}@di.ku.dk}}
}
\begin{document}
\maketitle

\begin{abstract}

Query augmentation is a crucial technique for refining semantically imprecise queries.
Traditionally, query augmentation relies on extracting information from initially retrieved, potentially relevant documents.
If the quality of the initially retrieved documents is low, then the effectiveness of query augmentation would be limited as well.
We propose Brain-Aug, which enhances a query by incorporating semantic information decoded from brain signals. 
Brain-Aug generates the continuation of the original query with a prompt constructed with brain signal information and a ranking-oriented inference approach. 
Experimental results on fMRI~(functional magnetic resonance imaging) datasets show that Brain-Aug produces semantically more accurate queries, leading to improved document ranking performance. 
Such improvement brought by brain signals is particularly notable for ambiguous queries.

\end{abstract}


\section{Introduction}

Understanding users' intentions is the key to the effectiveness of search engines. 
However, search engine users often struggle to precisely express their information needs, resulting in queries that are short~\cite{kacprzak2017query}, vague~\cite{yano2016quantifying,cronen2002predicting}, or inaccurately phrased, which compromise the retrieval effectiveness. 
To address this problem, query augmentation emerges as a crucial technique to refine the original queries into more effective expressions~\cite{38lavrenko2017relevance,mei2008query}. 
Traditionally, this reformulation process relies heavily on external document information such as expanding the query with contents from documents users have engaged with~\cite{chen2021hybrid,ahmad2019context,pereira2020iterative}.


The advent of neurophysiological interfaces offers a novel source of data to understand users' search intentions~\cite{ye2022towards,michalkova2024understanding}.
In \ac{IR} scenarios, several studies have revealed that brain signals can be used to predict users' relevance perception~\cite{ye2022don,eugster2014predicting,56pinkosova2020cortical} and cognitive state~\cite{51moshfeghi2016understanding}.
These advances open new avenues in using brain signals as an alternative to conventional signals for query augmentation.
Existing studies have investigated the use of brain signals to predict the relevance of perceived input~\citep{eugster2016natural}, which can be further used to extract relevant content for query augmentation~\citep{ye2022brain, ye2023relevance}.
The current process of query augmentation still relies on the quality of initially retrieved documents and cannot kick off before potentially unsatisfactory user interactions with those documents.
In this paper, we propose \emph{query augmentation with brain signals}~(Brain-Aug), which directly refines queries submitted by users through decoding semantics from their brain signals.
With the help of computational language models, Brain-Aug proposes two techniques to effectively refine queries:
(i)~Prompt construction with brain signals: Brain signals corresponding to the query context are decoded into the language model's latent space to construct prompts accordingly; 
(ii)~Training based on next token prediction and ranking-oriented inference: 
We teach the model to predict tokens in relevant documents as query continuation during training. 
Ranking-oriented features, i.e., \ac{IDF}, are incorporated to generate effective query continuation that can distinguish different documents during inference.

We conduct experiments on three \ac{fMRI} datasets.
Results show that Brain-Aug can accurately generate query continuations for its augmentation and improve the ranking performance.
Further investigation delves into different types of queries and shows that brain signals are particularly useful in enhancing the performance of ambiguous queries.


\begin{figure*}[t]
  \centering
  \includegraphics[width=1\linewidth]{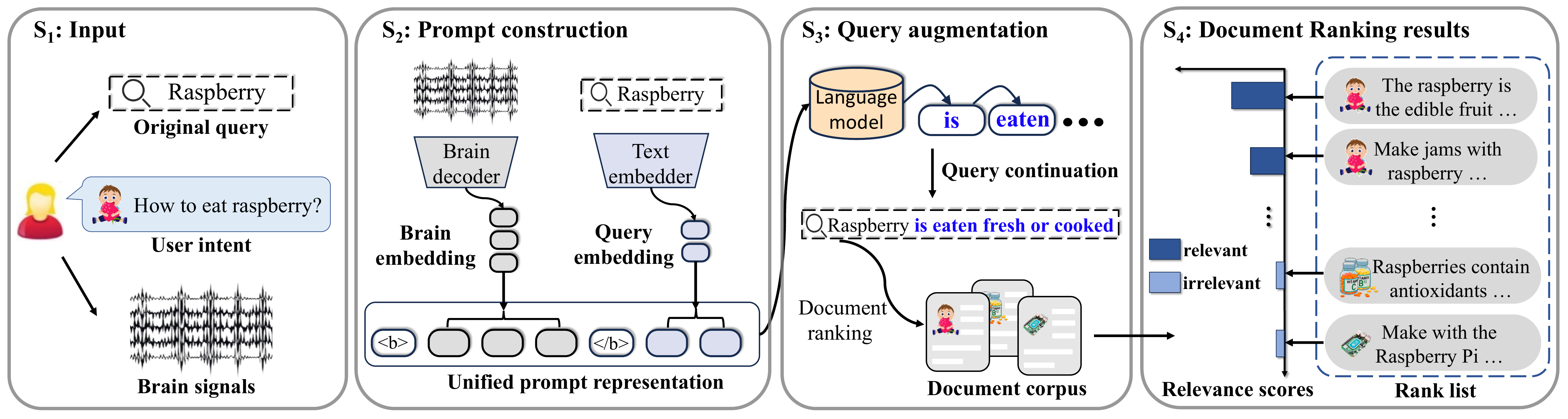}
  \caption{The procedure of query augmentation by decoding semantics from brain signals~(Brain-Aug). }
  \label{fig:procedure} 
\end{figure*}


\section{Related Work}
\textbf{Query augmentation.}
Traditionally, query augmentation can be categorized into two types: based on pseudo-relevance signals~\cite{bi2019iterative,38lavrenko2017relevance} and based on user signals~\cite{li2020systematic}. 
Approaches based on pseudo-relevance signals usually treat top-ranked documents in the initial retrieval step as relevant.
Based on these relevant documents, \citet{57rocchio1971relevance} and \citet{38lavrenko2017relevance} adopt a vector space model and a language model for refining the query representation to be closer to the top-ranked documents, respectively.
In contrast, approaches based on user signals usually integrate information from documents the user has previously interacted with or queries they submitted historically.
E.g., \citet{chen2021hybrid} and \citet{ahmad2019context} build a sequence model to extract semantic representations from historical clicked documents to refine the query representation.
Existing methods, either based on pseudo signals or user signals, are limited by their reliance on the quality of the documents and the accuracy of estimating their relevance.

\smallskip\noindent%
\textbf{Neuroscience \& IR.}
There is increasing literature that adopts neuroscientific methods into IR scenarios~\cite{11chen2022web,25gwizdka2017temporal,52mostafa2016deepening}.
For example, \citet{11chen2022web} built a prototype in which users can interact with the search systems with a brain-computer interface. 
\citet{3allegretti2015relevance, 51moshfeghi2016understanding, michalkova2024understanding} conducts a series of work to study the cognitive mechanisms involved in the process of information retrieval. 
A common finding observed by existing literature is that\citep{3allegretti2015relevance,eugster2014predicting} brain signals can be utilized to as a relevance indicator. 
This indicator can be employed for query rewriting~\cite{ye2022brain, eugster2016natural}.
Although this paradigm has been shown to be effective, it still relies on the quality of the retrieved documents.
On the other hand, other studies have demonstrated that semantics could be decoded to some extent with brain signals~\cite{wang2022open} such as \ac{fMRI}~\cite{xi2023unicorn,ye2023language,zou2021towards}
and \ac{MEG}~\cite{defossez2023decoding}.
However, there is currently a lack of research investigating the utilization of the decoded semantics for query augmentation.


\section{Method}


We first formalize the query augmentation task and then present Brain-Aug. 

\subsection{Task formalization}
The \emph{input} to the task of augmenting queries with brain signals is a query submitted by a user plus the brain signals associated with the query context. 
We use $Q$ to denote the query that is composed of $n$ tokens, $Q=\{q_1,q_2,\ldots,q_n\}$. 
We use $B=\{b_1, \ldots, b_t\} \in \mathbb{R}^{t\times c}$ to represent the brain signal, which is a sequence of features extracted from \ac{fMRI} data, where $c$ is the number of \ac{fMRI} features and $t$ is the number of time frames in which brain recordings are collected. 

Given the input query and brain signals, the \emph{task} is to learn an autoregressive function $F$ to refine the query based on the user's cognitive process. 
$F$ generates a query continuation $M=\{m_1,..., m_k\}$, which will be concatenated to the initial query $Q$ as the augmentated query. 
Let $m_i$ be the $i$-th token in $M$, the generation process is formalized as:
\begin{equation}
\mbox{}\hspace*{-1mm}
m_i = F(\{q_1, \ldots, q_n, m_1, \ldots, m_{i-1}\}, B; \Theta),    
\end{equation}
where $\Theta$ is the model parameters of $F$.

The effectiveness of query augmentation is measured \emph{extrinsically} using the document ranking performance.  
Formally, let $\mathcal{D}$ be a document corpus and $G$ be a ranking model~(e.g., BM25~\cite{robertson2009probabilistic}, RepLLaMA~\cite{ma2023fine}). 
The ranking model $G$ estimates a ranking score $G(\{Q, M\}, d)$ for each document $d \in \mathcal{D}$ and the document ranking performance can be measured by a ranking-based metric such as \ac{NDCG}~\cite{29jarvelin2002cumulated} or \ac{MAP}~\cite{30jarvelin2017ir}.

\zjtmerged{
}

\subsection{Overall procedure}


Fig.~\ref{fig:procedure} provides an overview of the four-stage process of Brain-Aug:
\begin{enumerate*}[label=$S_{\arabic*}:$,nosep]
    \item Input to Brain-Aug consists of the original query and brain signals associated with the user's cognitive response within the query context.
    \item Then a brain decoder is trained to align the representations of brain signals with the representation space of text embedding in the language model. This allows for creating a unified prompt representation that jointly models the brain responses and original queries.
    \item A language model is adopted to generate the continuation of the original query by using a unified prompt representation. A ranking-oriented inference method is utilized to enhance the generation process to improve the ranking performance.
    \item In this case, the original query ``Raspberry'' (sampled from Pereira's dataset in our experiment) is augmented to ``Raspberry is eaten fresh or cooked''. Consequently, documents with a focus on the subtopic of ``eating raspberry'' are ranked higher than those on ``raspberry's nutrition'' or ``raspberry Pi''.
\end{enumerate*}


\subsection{Prompt construction}

Motivated by existing literature that combines multimodal information as prompt~\cite{ye2023language,liu2023visual}, the prompt for Brain-Aug is constructed by integrating the textual query with cognitive information derived from brain signals.
First, the query's text $Q$ is directly fed to the language model's embedding layer $f_q$ to transform the tokens into latent vectors $V^{Q}=\{v^q_1, \ldots, v^q_i, \ldots, v^q_n\} \in \mathbb{R}^{n\times d}$, where $n$ is the number of tokens, $d$ is the embedding size of the language model.

Second, a brain decoder $f_b$ is devised to embed each brain representation $b_i \in B$ into the same latent space $\mathbb{R}^d$, which can be formulated as $v^B_i=f_b(b_i)$.
Based on preliminary empirical comparisons of transformers~\cite{vaswani2017attention}, linear layer, \ac{MLP}, and \ac{RNN}, we decide to construct the brain decoder as a deep neural network $f_b$ comprises 
\begin{enumerate*}[label=(\roman*)]
\item a MLP network $f_m$ with ReLU~\cite{fukushima1980neocognitron} as the activation function, and 
\item a position embedding $P=\{p_1, \ldots, p_t\} \in \mathbb{R}^{t \times c}$.
\end{enumerate*}
The position embedding is initialized using a uniform distribution. 
Element-wise addition is applied where each position embedding $p_i \in P$ is added to its corresponding \ac{fMRI} features $b_i \in B$.
The multi-layer perceptron network $f_m$ is constructed with an input layer and two hidden layers that have the same dimensionality $c$ as the input \ac{fMRI} features, as well as the output layer with the dimensionality of $d$.
In summary, the \ac{fMRI} features corresponding to the $i$-th time frame, i.e., $b_i$, are fed into the brain decoder $f_b$, which can be expressed as:
\begin{equation}
    v^B_i=f_b(b_i)=f_{mlp}(p_i+b_i).
\end{equation}
Finally, the brain embedding $V^{B}$ and the query embedding $V^{Q}$ are concatenated with embeddings of two special tokens, i.e., $\langle b\rangle$ and $\langle/b\rangle$, marking the beginning and end of the brain embedding, respectively. 
The two special tokens are randomly initialized as one-dimensional vectors aligned with the dimensional structure of token embeddings in the language model. 
As a result, the prompt sequence $S$ can be represented as:
\begin{equation}
    S=\{\langle b\rangle,v^B_1,\ldots,v^B_t,\langle/b\rangle,v^W_1, \ldots, v^W_n\}.
\end{equation}
This sequence, integrating both brain information and textual data, can be input to the language model for generating the query continuation.

Prior to the main training task detailed in Section~\ref{sec:Training objective}, a warmup step~\cite{huang2023language} is adopted to align the distribution of the brain embedding with that of the text token's embeddings, ensuring that the brain embedding is primed for integration with the text prompt embedding. 
To streamline the process and enable training in an unsupervised manner, each $v^B_i \in V^B$ is mapped to the mean value of the corresponding query embeddings, i.e., $\frac{1}{n}\sum_{j=1}^n v^Q_j$. 
\Ac{MSE} loss is adopted for the warmup process:
\begin{equation}
\textstyle
    L_\mathit{MSE} = \frac{1}{t} \sum_{i=1}^{t} \left(v^B_i - \frac{1}{n}\sum_{j=1}^{n}v^Q_j \right)^2.
\end{equation} 

\subsection{Training objective}
\label{sec:Training objective}
Given the unified prompt $S$, the training task is selected as the next token prediction task which predicts the continuation of $S$.
%
The prompt sequence $S$ is fed into a language model, e.g., the 7B version of LLaMA~\cite{touvron2023llama} in our implementation. 
The language model then estimates the likelihood of the ground truth continuation $M^{*}=\{m^{*}_1,\ldots,m^{*}_k\}$ by using an autoregressive function $P_{\text{LM}}(m^{*}_{i}\mid \{m^{*}_1,\ldots,m^{*}_{i-1}\},S)$ over the sequence $S$. 
The training objective is to maximize the likelihood of generating the ground truth continuation:
\begin{equation}
\small
\mbox{}\hspace*{-2mm}
    \mathop{\max}_{\Theta} \!= \!\sum_{i=1}^k \!\log(P_{\text{LM}}(m^{*}_i \!\mid\! \{m^{*}_1, \ldots, m^{*}_{i-1}\},S;\Theta)),
\hspace*{-2mm}\mbox{}
\end{equation}
where $\Theta=\{\Theta^{LM}, \Theta^{f_b}, \Theta^{sp}\}$ is the model parameters, $\Theta^{LLM}$, $\Theta^{f_b}$, and $\Theta^{sp}$ are the parameters of the language model, the brain decoder, and the special tokens $\langle b\rangle$ and $\langle/b\rangle$, respectively.

Here, we propose to set the ground-truth label of the continuation as the content from the labeled relevant documents~(see Section~\ref{Experimental setup} for details). 
First, when a document is relevant, it must contain important information and tokens that can potentially be decoded from the brain signals~\cite{pereira2018toward}. 
Second, teaching models to expand queries with terms in potentially relevant documents could improve the performance of downstream retrieval models~\cite{robertson2009probabilistic}.
The training process follows the ``prompt tuning'' approach~\cite{liu2023gpt} by keeping the parameters of the language model unchanged and fine-tuning only the prompt representation, i.e., $\Theta^{f_b}$, and $\Theta^{sp}$. 
In this way, we can train Brain-Aug efficiently with limited training data.

\subsection{Ranking-oriented inference}
During the inference stage, the generated continuations should also be able to distinguish between different documents.
Therefore, we incorporate the \ac{IDF} information~\cite{robertson2004understanding} of each token in the vocabulary when generating query continuation $\hat{M} = \{\hat{m}_1, \ldots,\hat{m}_k\}$.
Let $\text{IDF}(\hat{m})$ be the \ac{IDF} of token $\hat{m}$, then the generation likelihood of each token in $\hat{m}_i \in \hat{M}$ during the inference stage can be estimated as:  
\begin{equation}
\small
\mbox{}\hspace*{-2mm}
   P_{\text{inf}}(\hat{m}_i) \!=\! \frac{P_\text{LM}(\hat{m}_i) \!+\! \alpha\, \text{IDF}(\hat{m}_i)}{\sum_{m \in \text{Vocab}}(P_\text{LM}(m) \!+\! \alpha\, \text{IDF}(m))},
\hspace*{-1mm}\mbox{}
\end{equation}
where {\small $P_{LM}(m)=P_{LM}(m\mid \{\hat{m}_{1}, \ldots,\hat{m}_{i-1}\},S;\Theta)$} represents the estimated likelihood of the next token $m$ given the previously generated tokens $\{\hat{m}_{1}, \ldots,\hat{m}_{i-1}\}$, $\alpha$ is a hyperparameter, Vocab indicates the language model's vocabulary. 
This approach ensures that the query's continuation is not only contextually relevant but also effective in distinguishing documents in the retrieval process.


\section{Experimental Setup}
\label{Experimental setup}


Next, we detail our experimental settings, which are designed to address three research questions:
\begin{enumerate*}[label=\textbf{(RQ\arabic*)}]
    \item Is it possible to generate an augmented query with user's brain signals?
    \item Can we improve document ranking performance using the augmented query?
    \item How do brain signals improve different queries for document ranking?
\end{enumerate*}
Together, these questions help us to understand the effectiveness of Brain-Aug to refine a query and improve ranking performance.
Below, we describe the datasets and baselines.
More implementation details are provided in Section~\ref{sec:implementation details}.



\subsection{Datasets}
Three publicly available fMRI datasets are adopted, namely 
Pereira's dataset~\cite{pereira2018toward},
Huth's dataset~\cite{lebel2023natural}, and the 
Narratives dataset~\cite{nastase2021narratives}. 
We process the text stimuli in these datasets to transform them into ranking datasets consists of a document corpus and a set of queries. 
The dataset information is provided in Section~\ref{sec:Dataset statistics}.

\subsection{Data processing}
Due to the lack of clear definitions for query and document parts in those existing fMRI datasets, we use the \ac{ICT} setting~\cite{izacard2021unsupervised,lee2019latent} to test the query augmentation performance.
The \ac{ICT} setting selects a text span in the document as a pseudo query and the corresponding document is treated as relevant for this query.
Formally, for a document $D=\{w_1,\ldots,w_m\}$, ICT extracts a span $Q=\{w_l, w_{l+1},\ldots,w_r\}$ to form a relevant query-document pair $\{Q,D\backslash Q\}$, where $D\backslash Q=\{w_1,\ldots,w_{l-1},w_{r+1},\ldots,w_m\}$.

In Pereira's dataset, each document consists of 3-4 sentences, which are presented to the user as visual stimuli one by one.
Due to the length of a sentence being too long as a query, we truncate the first one-third and two-thirds of the sentence to construct two queries for each sentence, resulting in 6-8 relevant query-document pair for each document.
In Huth's and Narratives datasets, continuous contents are presented to the user as auditory stimuli.
We utilize a fixed time interval of 20 seconds, which corresponds to 10 \ac{fMRI} scans, to segment the stimuli into documents. 
Then, smaller time intervals of 2, 4, and 6 seconds are employed to segment queries of varying lengths from the document.
We provide more details and statistical data for the document corpus and queries constructed in each dataset in Section~\ref{sec:Dataset preprocessing}.

Due to the variability in brain data across participants, we trained separate models for each participant and evaluated Brain-Aug using a five-fold cross-validation on each participant's data.
The data samples are randomly split into five folds according to which document they belong to.
Each fold of the cross-validation involves selecting one fold of the data as the test set, while the remaining four folds are split into training and validation sets.
The sizes of the training, validation, and testing sets were roughly proportional to 3:1:1, respectively.

\subsection{Training and evaluation setup}
We train Brain-Aug with a next token prediction task.
A data sample during this task consists of the query, its ground truth continuation, and corresponding brain signals.
The ground truth continuation is selected as the textual content presented within a fixed period of time after the query~(see Section~\ref{sec:Dataset preprocessing} for details).
Taking into account the delayed effect of \ac{fMRI} signals\cite{mitchell2008predicting}, we collect user's brain signals in a period of several seconds after the user perceives the textual content of the query. 
During this period, the user's brain representation has the potential to encode semantic information related to the query itself, as well as its continuation.

We first conduct \emph{query generation analysis} to investigate the ability of Brain-Aug to generate query continuation that matches the ground truth label.
The logarithm perplexity~\cite{meister2021language} is used to measure the likelihood of generating the ground truth continuation.
The lower perplexity indicates the language model deems the ground truth continuation as more expected.
We also investigate language similarity to demonstrate the extent to which the generated continuation is similar to the ground truth using the Rouge score~\cite{lin2004rouge}.

Next, we augment the original query with its generated continuation and evaluate its performance in terms of \textit{document ranking}. 
We employ document ranking metrics, including \ac{NDCG} at different cutoffs (10 and 20)~\cite{29jarvelin2002cumulated}, Recall@20, and MAP~\cite{30jarvelin2017ir}.

\subsection{Baselines and controls}
Given the augmented query, we select two ranking models for document ranking, i.e., a sparse ranking model, \textbf{BM25}~\cite{robertson2009probabilistic}, and a dense ranking model, \textbf{RepLLaMA}~\cite{ma2023fine}.
To assess whether Brain-Aug helps document ranking, we compare its document ranking performance with several \emph{baselines} and \emph{controls}.

As \emph{baselines} we select 
(i)~\textbf{the original query}, and 
(ii)~the query augmented with pseudo-relevance signals~(denoted as \textbf{Unsup-Aug}).
When using BM25 as the ranking model, we implemented RM3~\cite{38lavrenko2017relevance} as Unsup-Aug, which expands the query by selecting relevant terms from the top-ranked documents in the initial retrieval.
When using RepLLaMA as the ranking model, we implement Rocchio~\cite{bi2019iterative} as Unsup-Aug, which refines the query vector to be closer to the top-ranked documents.
(iii)~We also reported the additional results by first using Brain-Aug, followed by Unsup-Aug, denoted as \textbf{Brain+Unsup}.

As \emph{controls} we select variants or ablations of Brain-Aug.
The first control is Brain-Aug without any brain input~(denoted as \textbf{w/o Brain}), and thus the query continuation is generated solely depending on the original query and the language model.
The second control is Brain-Aug with randomly sampled brain input~(denoted as \textbf{RS Brain}).  
RS Brain involves sampling brain input that does not correspond to the query but is randomly selected from the same dataset.
The last control is Brain-Aug without ranking-oriented generation in which the generation likelihood of each token is estimated without the IDF weight~(denoted as \textbf{w/o IDF}). 


\section{Experiments and Results}
We first analyze the performance of the generated query continuation by comparing it with the ground truth label. 
Then we investigate the document ranking performance with Brain-Aug and examine the relationship between query features and their ranking performance.

\subsection{Query generation analysis}
\label{sec:Query generation analysis}
The query generation analysis results are presented in Table~\ref{tab:generation_performance}.
From Table~\ref{tab:generation_performance}, we have the following observations.

(1)~Brain-Aug exhibits lower perplexity and higher Rouge-L than its ablations without brain input~(w/o Brain) and randomly sampled brain signals as input~(RS Brain).
This indicates that the semantic information decoded from brain signals can be integrated with a query to construct a more effective prompt for generating query continuation. 

(2)~The overall perplexity and Rouge-L on the Pereira dataset are lower and higher than on the other two datasets, respectively.
This implies that the Pereira dataset, derived from Wikipedia data, exhibits superior performance in the task of query generation compared to the other two datasets, which are based on spoken stories.

(3)~The RS Brain outperforms w/o Brain across three datasets.
Although RS Brain uses brain signals that do not correspond to the current query context, the unified prompt can enable generating content that aligns with the common data distribution of language usage in the dataset~(e.g., all stimuli in Pereira's dataset are Wikipedia-style).
On other other hand, w/o Brain is equivalent to a standard language model that generates continuations soly based on the query text.
This difference explains RS Brain's superior performance compared to the w/o Brain.
However, in the discussion in Section~\ref{sec:Document ranking performance}, we will show that this performance improvement in query generation does not necessarily lead to an improvement in document ranking.

\begin{table}[t]
\centering
\small
\setlength{\tabcolsep}{1.6mm}
\begin{tabular}{@{} l l@{}c@{~}c @{}}
\toprule
\multicolumn{1}{c}{\textbf{Dataset}} & \multicolumn{1}{c}{\textbf{Query}} & \multicolumn{1}{c}{\textbf{$\log$(PPL)($\downarrow$)}} & \multicolumn{1}{c}{\textbf{Rouge-L($\uparrow$)}}   \\
\midrule
\multicolumn{1}{c}{\multirow{3}{*}{{{{Pereira's}}}}}    
        & w/o Brain & $2.219^*$   & $0.213^*$ \\
        & RS Brain & $1.967^*$   & $0.267^*$ \\
        & Brain-Aug & \textbf{1.946}\phantom{$^*$}   & \textbf{0.272}\phantom{$^*$} \\
        \midrule
\multicolumn{1}{c}{\multirow{3}{*}{Huth's}}      
        & w/o Brain & $3.573^*$   & $0.148^*$  \\
        & RS Brain & $3.111^*$    & $0.159^*$ \\
        & Brain-Aug & \textbf{2.997}\phantom{$^*$}   & \textbf{0.167}\phantom{$^*$} \\
        \midrule
\multicolumn{1}{c}{\multirow{3}{*}{{{{Narratives}}}}} 
        & w/o Brain & $4.328^*$   & $0.083^*$ \\
        & RS Brain & $3.532^*$   & $0.105^*$ \\
        & Brain-Aug & \textbf{3.471}\phantom{$^*$}   & \textbf{0.109}\phantom{$^*$} \\
\bottomrule
\end{tabular}
\caption{Query generation performance averaged across participants in different datasets. Best results in boldface. * indicates $p\leq 0.05$ for the paired t-test of \textit{Brain-Aug~(Ours)} and the controls. PPL indicates perplexity. } 
\label{tab:generation_performance}
\end{table}

\noindent%
\textbf{Answer to \textbf{RQ1}.} 
The results show that queries augmented with semantics decoded from brain signals are more aligned with the content of the relevant document with the help of brain signals.

\subsection{Document ranking performance}
\label{sec:Document ranking performance}
\begin{table*}[t]
\small
\centering
\begin{tabular}{l @{~~} l llll llll}
\toprule
\multicolumn{1}{c}{\multirow{2}{*}{\textbf{Dataset}}} & \multicolumn{1}{c}{\multirow{2}{*}{\textbf{Query}}} & 
\multicolumn{4}{c}{\textbf{BM25}} & \multicolumn{4}{c}{\textbf{RepLLaMA}} 
\\
\cmidrule(r){3-6}
\cmidrule{7-10}
& & \multicolumn{1}{c}{\textbf{N@10}} & \textbf{N@20} & \textbf{R@20} & \textbf{MAP} & \textbf{N@10} & \textbf{N@20} & \textbf{R@20} & \textbf{MAP} \\
\midrule
\multicolumn{1}{c}{\multirow{4}{*}{{{{Pereira's}}}}} & original & $0.643^{*,\dagger}$ & $0.664^{*,\dagger}$ & $0.888^{*,\dagger}$  & $0.594^{*,\dagger}$ & 0.878 & $0.881^{*,\dagger}$ & $0.964^{*,\dagger}$ & 0.858 \\ 
& Unsup-Aug & $0.646^{*,\dagger}$ & $0.655^{*,\dagger}$ & $0.924^{*,\dagger}$ & $0.590^{*,\dagger}$ & $0.872^{*,\dagger}$ & $0.877^{*,\dagger}$ & $0.951^{*,\dagger}$ & 0.855 \\
& Brain-Aug & 0.671 & \textbf{0.691} & \textbf{0.941} & \textbf{0.618} & \textbf{0.883} & \textbf{0.887} & \textbf{0.980} & \textbf{0.859} \\
& Brain+Unsup & \textbf{0.673} & 0.686 & 0.936 & 0.615 & 0.878 & 0.882 & 0.975 & 0.853 \\
\midrule

\multicolumn{1}{c}{\multirow{4}{*}{{{{Huth's}}}}} & original & $0.297^{*,\dagger}$ & $0.326^{*,\dagger}$ & $0.536^{*,\dagger}$ & $0.264^{*,\dagger}$ & $0.299^{*,\dagger}$ & $0.328^{*,\dagger}$ & $0.520^{*,\dagger}$ & $0.275^{*,\dagger}$ \\ 
& Unsup-Aug & $0.291^{*,\dagger}$ & $0.320^{*,\dagger}$ & $0.575^{\dagger}$ & $0.259^{*,\dagger}$ & $0.302^{*,\dagger}$ & $0.333^{*,\dagger}$ & $0.537^{*,\dagger}$ & $0.276^{*,\dagger}$ \\
& Brain-Aug & 0.306 & 0.340 & $0.569^{\dagger}$ & \textbf{0.273} & \textbf{0.310} & \textbf{0.342} & 0.550 & \textbf{0.281} \\
& Brain+Unsup & \textbf{0.309} & \textbf{0.342} & \textbf{0.580} & 0.269 & 0.308 & 0.340 & \textbf{0.552} & 0.279\\
\midrule

\multicolumn{1}{c}{\multirow{4}{*}{{{{Narratives}}}}} & original & $0.419^{*,\dagger}$ & $0.434^{*,\dagger}$ & $0.629^{*,\dagger}$  & $0.355^{*,\dagger}$ & $0.413^{*,\dagger}$ & $0.426^{*,\dagger}$ & $0.611^{*,\dagger}$ & $0.351^{*,\dagger}$ \\ 
& Unsup-Aug & 0.440 & $0.452^{\dagger}$ & $0.670^{\dagger}$ & $0.367^{*,\dagger}$ & $0.416^{*,\dagger}$ & $0.431^{*,\dagger}$ & $0.629^{*,\dagger}$ & $0.356^{*,\dagger}$ \\
& Brain-Aug & 0.441 & 0.458 & 0.669 & \textbf{0.382} & 0.430 & \textbf{0.446} & 0.641 & \textbf{0.382} \\
& Brain+Unsup & \textbf{0.445} & \textbf{0.462} & \textbf{0.678} & \textbf{0.382} & \textbf{0.432} & \textbf{0.446} & \textbf{0.642} & 0.380\\
\bottomrule
\end{tabular}
\caption{Document ranking performance averaged across participants. Best results in boldface. $*$/$\dagger$ indicates Brain-Aug / Brain+Unsup significantly outperforms the baseline ($p\leq 0.05$, paired t-test), respectively.}
\label{tab:overallPerformance}
\end{table*}

\noindent \textbf{Overall performance.} 
Table~\ref{tab:overallPerformance} shows the document ranking performance with original queries, queries augmented with unsupervised signals~(Unsup-Aug), and queries augmented with brain signals~(Brain-Aug).
We observe:

(1)~Regardless of whether BM25 or RepLLaMa is used as the ranking model, Brain-Aug substantially outperforms the original query and Unsup-Aug.
The only exception is observed when using Rep\-LLaMa and metric MAP on Pereira's dataset.
A possible explanation for this exception is the Rep\-LLaMA's high performance on the Pereira dataset, which we discuss in observation (3).

(2)~When considering various datasets and metrics, the Unsup-Aug query does not consistently outperform the original query. 
Significant differences between the performance achieved by the Unsup-Aug query and the original query emerge on the metric of Recall@20 when using BM25 as the ranking model.
This observation suggests that Unsup-Aug, which improves query representation by tackling term mismatch issues, leads to an improvement in recall.
When Brain-Aug is combined with Unsup-Aug (Brain+Unsup), we observe a performance gain when compared to Unsup-Aug. 
This highlights the effectiveness of brain signals in query augmentation and underscores the potential of combining them with traditional signals.


(3)~We observe little difference in performance between RepLLaMa and BM25 on Huth's dataset and Narratives's dataset.
This implies that in a zero-shot setting and cross-domain scenario~(the datasets are derived from spoken stories, which differs from the training data of Rep\-LLaMa), dense retrieval models like Rep\-LLaMa are not necessarily better than traditional sparse retrieval models like BM25.
This phenomenon is also observed in the BEIR dataset~\citep{thakur2021beir}.
However, in Pereira's dataset, Rep\-LLaMa shows significant improvement over BM25 with different query inputs. 
The impressive performance of RepLLaMa on Pereira's dataset can likely be attributed to the fact that the data in Pereira are likely to be used in the original construction of Rep\-LLaMa.

\noindent\textbf{Decomposing Brain-Aug.}
Next, we investigate the contribution of brain signals and the ranking-oriented inference approach to Brain-Aug.
Experimental results are presented in Table~\ref{tab:ablation_performance}.
First, we observe that removing~(w/o Brain) or random sampling the brain inputs~(RS Brain) leads to a decrease in performance.
This indicates that semantic information decoded from brain signals within the query context enhances the query.
Furthermore, while RS Brain consistently outperforms w/o Brain approach in terms of generation perplexity~(see Section~\ref{sec:Query generation analysis}), it struggles to achieve better document ranking performance on the Huth's and Narratives datasets. 
This can be attributed to the fact that RS Brain, despite generating content that closely matches the token distribution of the whole dataset and reducing perplexity, fails to effectively differentiate between different documents within the dataset without semantics related to the query context.
Last, we also observe a significant performance improvement when comparing Brain-Aug against its ablation without ranking-orient generation~(w/o IDF).
This suggests the importance of generating content that can be used to differentiate between documents.

\begin{table}[t]
\small
\centering
\begin{tabular}{l l  cc}
\toprule
\multicolumn{1}{c}{\textbf{Dataset}} & \multicolumn{1}{c}{{\textbf{Query}}} & {\textbf{NDCG@20}}& {\textbf{MAP}}   
\\
\midrule
\multicolumn{1}{c}{\multirow{4}{*}{{Pereira's}}}    & w/o Brain & $0.665^*$   & $0.586^*$ \\
                            & RS Brain & $0.678^*$ & $0.604^*$ \\
                            & w/o IDF   & $0.684^*$   & $0.609^*$ \\
                            & Brain-Aug & \textbf{0.691}\phantom{$^*$}   & \textbf{0.618}\phantom{$^*$} \\\midrule 
\multicolumn{1}{c}{\multirow{4}{*}{{{{Huth's}}}}}      & w/o Brain & $0.332^*$   & $0.265^*$  \\
                            & RS Brain & $0.321^*$    & $0.256^*$ \\
                            & w/o IDF   & $0.332^*$   & $0.266^*$ \\
                            & Brain-Aug & \textbf{0.340}\phantom{$^*$}    & \textbf{0.273}\phantom{$^*$}  \\\midrule
\multicolumn{1}{c}{\multirow{4}{*}{{{{Narratives}}}}} & w/o Brain & $0.452^*$   & $0.368^*$ \\
                            & RS Brain & $0.448^*$   & $0.367^*$ \\
                            & w/o IDF   & $0.450^*$    & $0.373^*$ \\
                            & Brain-Aug & \textbf{0.458}\phantom{$^*$}    & \textbf{0.382}\phantom{$^*$}  \\\bottomrule
\end{tabular}
\caption{Document ranking performance of \textit{Brain-Aug~(ours)} and its controls with ranking model BM25. Best results in boldface. * indicates $p\leq 0.05$ for the paired t-test of \textit{Brain-Aug} and the baseline.}
\label{tab:ablation_performance}
\end{table}

\begin{figure}[t]
  \centering
  \begin{subfigure}{0.45\columnwidth}
    \includegraphics[width=\linewidth]{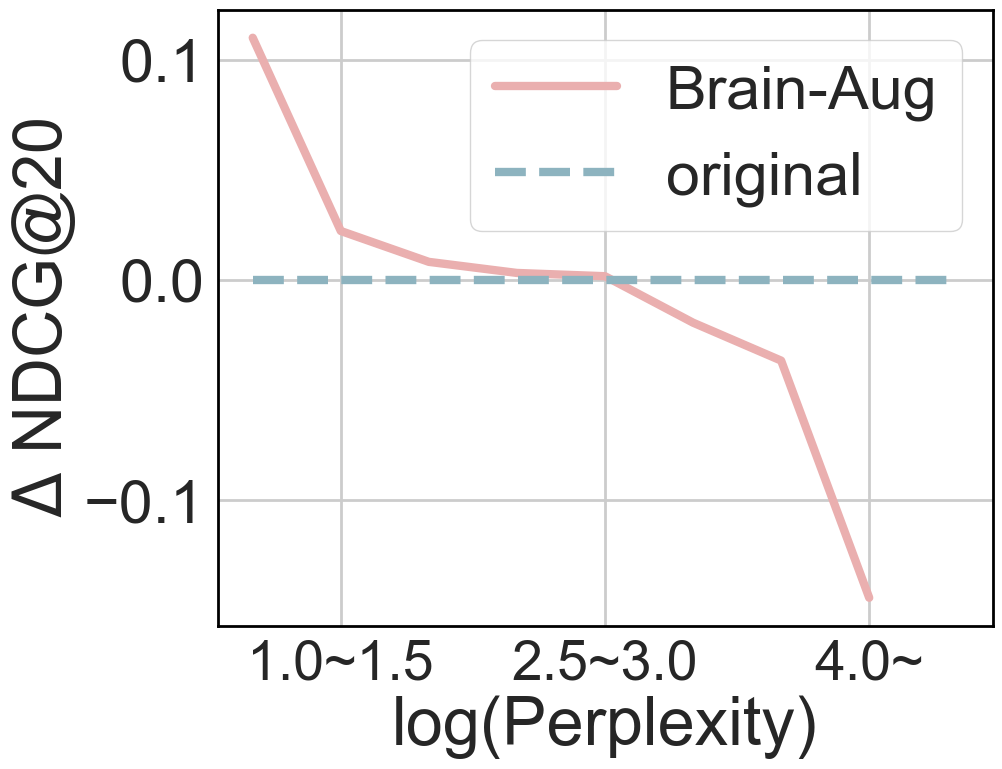}

    \vspace*{-2mm}
    \caption{Brain-Aug vs.\ original}
    \label{fig:relationship_original}
  \end{subfigure}
  \hfill
  \begin{subfigure}{0.47\columnwidth}
    \includegraphics[width=\linewidth]{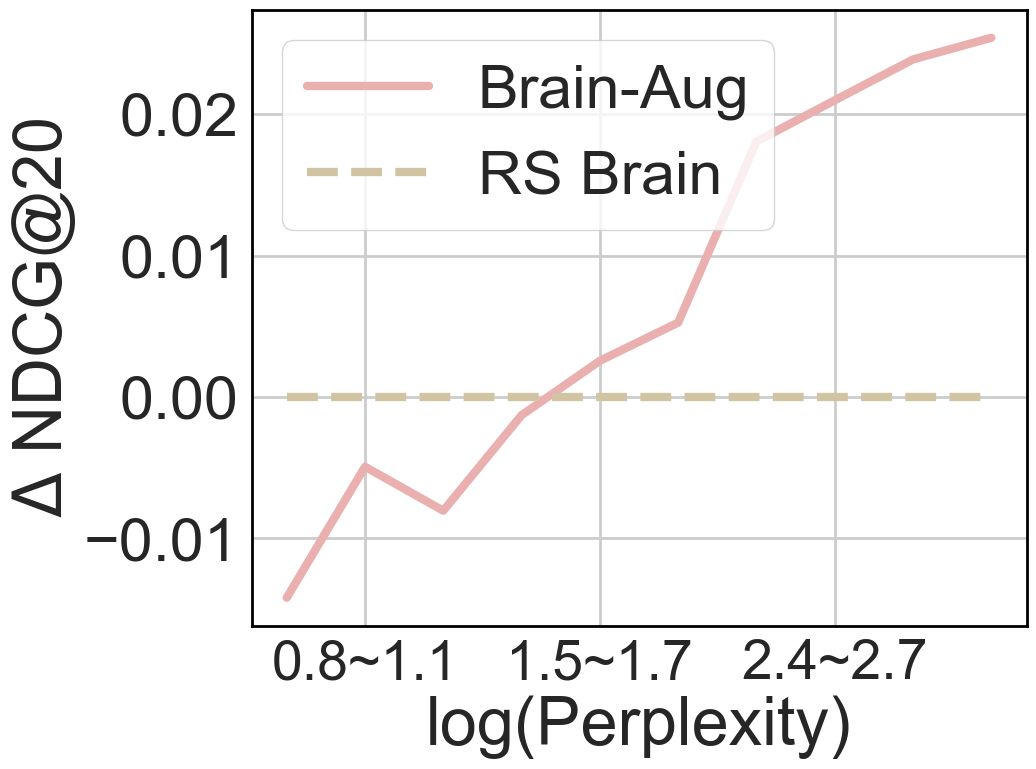}

    \vspace*{-2mm}    
    \caption{Brain-Aug vs.\ RS B}
    \label{fig:relationship_RS}
  \end{subfigure}
  \caption{Relationship between document ranking performance and perplexity of ground-truth query continuation in Pereira's dataset. ``RS B'' indicates the ablation of Brain-Aug that randomizes brain inputs. $\Delta$ NDCG@20 indicates performance gains of Brain-Aug. }
  \label{fig:relationship}
\end{figure}

\noindent \textbf{Relationship between document ranking and query generation performance.} 
Fig.~\ref{fig:relationship} illustrates the relationship between the document ranking performance of Brain-Aug and RS Brain and the perplexity of query continuation measured using RS Brain.
The lower perplexity of query generation indicates a higher likelihood of generating more accurate query continuation. 
This higher likelihood, as shown in Fig.~\ref{fig:relationship_original}, further leads to an increase in document ranking performance.
Conversely, Fig.~\ref{fig:relationship_RS} shows a different trend: when the perplexity is higher, the performance gain of Brain-Aug with its ablation RS Brain is higher.
This implies that when generating accurate query continuations is difficult, semantics decoded from the query context with brain signals is more beneficial.
This observation is consistent with findings by \citet{ye2023language} that the addition of brain signals lead to a more substantial performance improvement when generating continuations with higher uncertainty .

\begin{table*}[t]
    \centering
    \small
    \begin{tabular}{lm{0.2\textwidth} m{0.5\textwidth}c}
    \toprule
    \multicolumn{1}{c}{\textbf{Method}} & \multicolumn{1}{c}{\textbf{Query Content}} & \multicolumn{1}{c}{\textbf{Top-ranked document}} & \textbf{Relevance} \\
    \midrule
    \multicolumn{1}{c}{Original} & The \textcolor{true_blue}{ shaking can } & $d_{21}$: 
 The wind from the hurricane \textcolor{true_blue}{shook} the house, shattering a window ... Later that night, with the wind \textcolor{true_blue}{shaking} the house, ... & 0 \\
    \midrule
    \multicolumn{1}{c}{Unsup-Aug} & The \textcolor{true_blue}{shaking can} \textcolor{myPurple}{from house wind} & $d_{21}$: The \textcolor{myPurple}{wind} \textcolor{myPurple}{from} the hurricane \textcolor{true_blue}{shook} the \textcolor{myPurple}{house}, shattering a window ... Later that night, with the \textcolor{myPurple}{wind} \textcolor{true_blue}{shaking} the house ... & 0 \\
    \midrule
    \multicolumn{1}{c}{RS Brain} & The \textcolor{true_blue}{shaking can} \textcolor{myPurple}{ last anywhere from a few seconds to several minutes} & $d_{21}$: The wind \textcolor{myPurple}{from} the hurricane \textcolor{true_blue}{shook} the house, shattering a window in the kitchen. ... Later that night, with the wind \textcolor{true_blue}{shaking} the house, we fell asleep huddled on the sofa. & 0 \\
    \midrule
    \multicolumn{1}{c}{Brain-Aug} & The \textcolor{true_blue}{ shaking can} \textcolor{myPurple}{ be caused by an earthquake} & $d_{13}$: \textcolor{myPurple}{Earthquakes} \textcolor{true_blue}{shake} the ground and \textcolor{true_blue}{can} knock down buildings and other structures. \textit{[MASK]} also trigger landslides and volcanic activity. Most \textcolor{myPurple}{earthquakes} are \textcolor{myPurple}{caused by} ... & 1 \\
    \bottomrule
    \end{tabular}
\caption{Examples of document ranking with BM25 using the original query or the augmented query in Pereira's dataset. Text in \textcolor{true_blue}{blue} and in \textcolor{myPurple}{purple} indicates content in the original query and generated by the query augmentation method, respectively. \textit{[MASK]} indicates the position of the query ``The shaking can'' in the ICT setting. }
\label{tab:case study}    
\end{table*}

\noindent \textbf{Example cases.}
Table~\ref{tab:case study} presents example cases with the original query ``The shaking can'' which is sampled from document $d_{13}$ in Pereira's dataset. 
Brain-Aug leverages brain signals to expand the query with ``be caused by an earthquake''. 
As a result, the relevant document with the topic of the earthquake, $d_{13}$, is appropriately ranked at the top of the search results.
Example cases for Huth's and Narratives dataset are provided in Section~\ref{sec:appendix_example}.

\begin{figure*}[t]
  \centering
  \begin{subfigure}{0.50\columnwidth}
    \includegraphics[width=\linewidth]{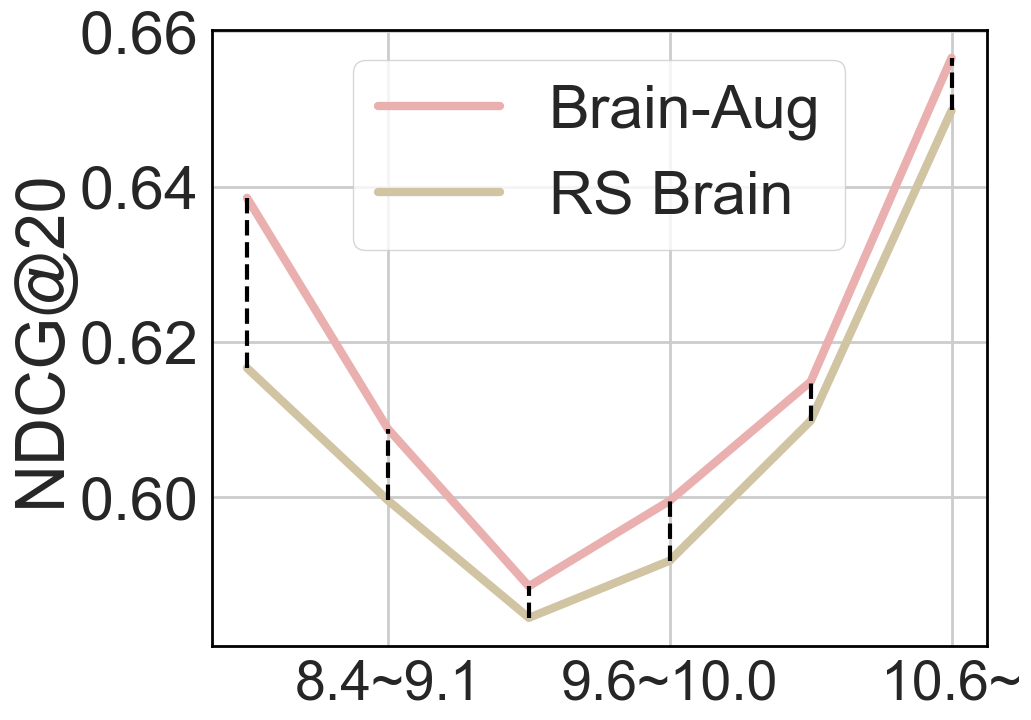}

    \vspace*{-2mm}
    \caption{Avg ICTF}
    \label{fig:qpp_ictf}
  \end{subfigure}
  \hfill
  \begin{subfigure}{0.50\columnwidth}
    \includegraphics[width=\linewidth]{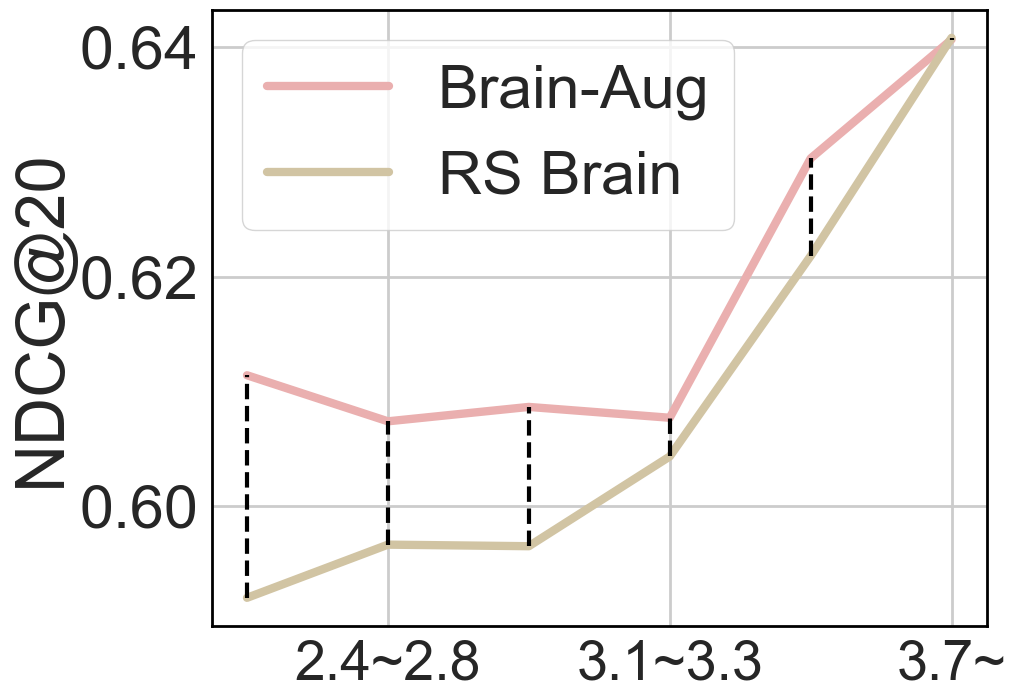}

    \vspace*{-2mm}
    \caption{Avg IDF}
    \label{fig:qpp_idf}
  \end{subfigure}
  \hfill
  \begin{subfigure}{0.50\columnwidth}
    \includegraphics[width=\linewidth]{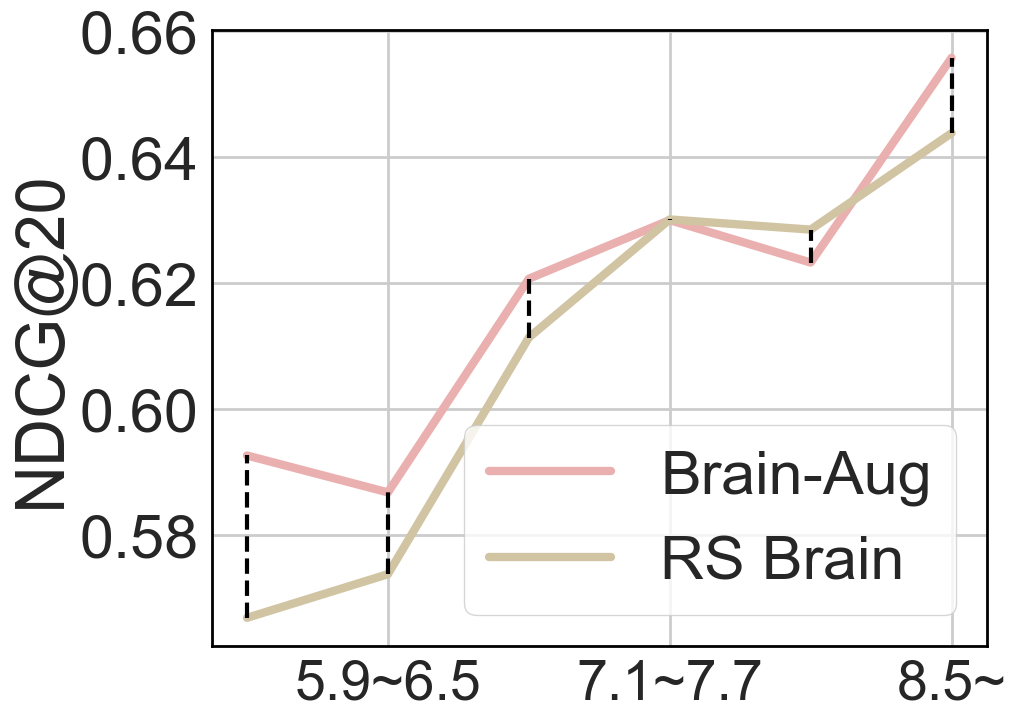}

    \vspace*{-2mm}
    \caption{Specificity}
    \label{fig:qpp_specificity}
  \end{subfigure}
  \hfill
  \begin{subfigure}{0.50\columnwidth}
    \includegraphics[width=\linewidth]{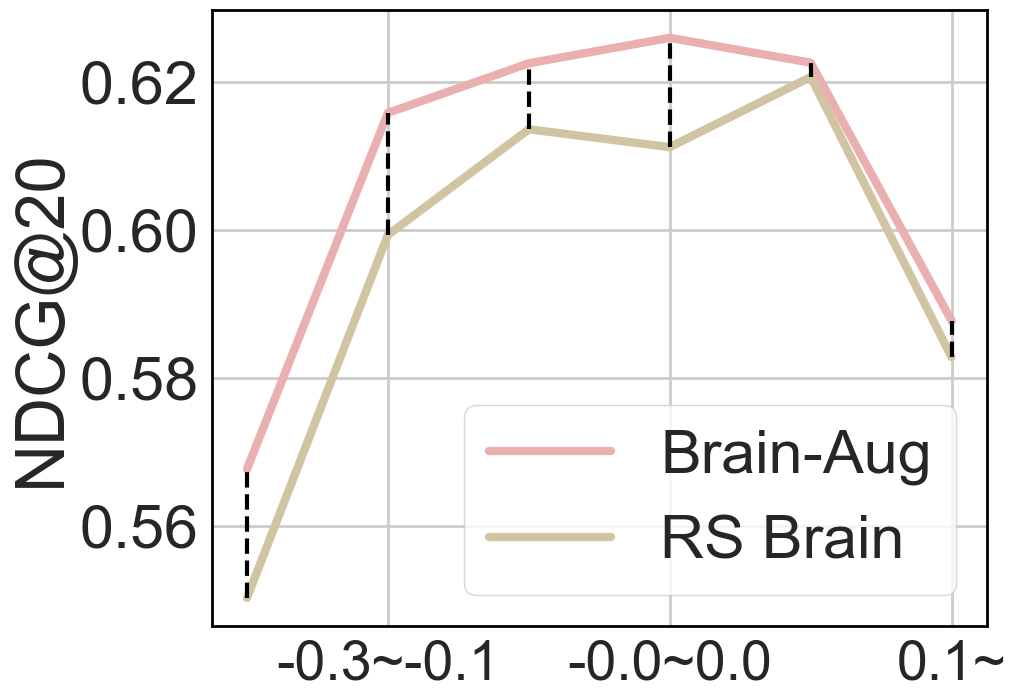}

    \vspace*{-2mm}
    \caption{Clarify}
    \label{fig:qpp_clarify}
  \end{subfigure}

    \vspace*{-2mm}
  \caption{Document ranking performance w.r.t.\ different query features in Pereira's dataset.}
  \label{fig:qpp}
\end{figure*}

\noindent%
\textbf{Answer to \textbf{RQ2}.}
We verified that a query augmented with semantics decoded from brain signals can significantly enhance document ranking performance. 
This performance enhancement is more pronounced when the generated query continuation is more accurately aligned with the query context.

\subsection{Query performance analysis}
Next, we investigate the performance improvement achieved by Brain-Aug for different queries by grouping queries according to their features.
We select four query features: three pre-retrieval features~(calculated based on query tokens), i.e., \textit{ICTF}, \textit{IDF}, and \textit{specificity} score~\cite{shtok2012predicting}, and one post-retrieval feature~(calculated based on the information of retrieved documents), i.e., \textit{clarify}  score~\cite{cronen2002predicting,meng2023query}.
For details on the query features, see Section~\ref{sec:Query performance features}.
We conjecture that larger feature values correspond to a more clarified query and usually result in better retrieval quality.

Fig.~\ref{fig:qpp} depicts the document ranking performance w.r.t.\ different query features on Pereira's dataset.
We have two key observations.
(i)~When the averaged IDF, specificity score, and clarity score increase, both Brain-Aug and the RS Brain show an improvement in retrieval performance. 
This indicates that a more specific query usually has a better retrieval performance.
(ii)~The performance gain of Brain-Aug compared to RS Brain is more pronounced when these features experience a decrease. 
This observation is supported by a significant negative Pearson's $r$ between the improvement in NDCG@20 for Brain-Aug compared to RS Brain and the averaged ICTF, averaged IDF, specificity score, and clarity score, which are $-0.14$, $-0.19$, $-0.17$, and $-0.32$, respectively.
This indicates that the performance improvement brought by brain signals is larger in queries prone to be vague or ambiguous.

\noindent%
\textbf{Answer to \textbf{RQ3}.}
We have observed that queries prone to ambiguity~(e.g., containing tokens with lower IDF scores or with low clarify scores) stand to gain more from Brain-Aug.

\section{Discussion and Conclusion}
Existing research incorporating physiological signals in IR tasks, whether based on eye-tracking~\cite{bhattacharya2020relevance} or brain signals~\cite{ye2023relevance,eugster2014predicting}, has relied on predicting relevance of presented information.
Here, we have investigated an alternative approach for directly augmenting queries  based on the semantic information decoded from fMRI brain signals.
Our findings revealed that decoding semantic representations from brain signals can enhance the generation of queries and subsequently improving document ranking. 
Moreover, we have observed that brain signals are more effective when the content to be generated has higher perplexity, indicating that decoded semantic information for unlikely query augmentations is more effective than it is for likely query augmentations. 
In conclusion, our findings open a horizon for new types of methods for understanding users by decoding semantics associated with information needs directly from brain signals.
This process can kick off naturally as it happens as part of perceiving information and without requiring users to engage with any particular interaction technique or user interface.

\clearpage
\section{Limitations}
Our work has the following limitations pointing towards promising avenues for future research:
(i)~Our study utilized fMRI signals, which are not readily accessible in real-world human-computer interaction scenarios and have a significant delay of 2-8 seconds. 
More commonly used signals, such as \ac{EEG}, have lower signal-to-noise ratios, which may limit their utility for semantic decoding.
Currently, there is a lack of evidence that EEG can effectively decode semantics.
In recent years, sensor technology like \ac{fNIRS} and \ac{MEG} may become promising directions for future research. 
(ii)~Our experiments simulate the document ranking with an ICT setting and show significant improvements over the baselines and carefully designed controls. 
Although ICT is commonly used to test retrieval performance, it is different from the most realistic search interaction. 
This simulation with ICT was driven by its advantage in building a sufficient number of queries and obtaining the corresponding query context to construct a substantial amount of training data.
In the future, it would be worthwhile to explore settings that closely resemble real-world query interaction. This can be done through approaches such as training with ICT and testing with another corpus of queries, or by designing few-shot learning or cross-subject training models to enable query augmentation with a limited amount of data.

\section{Ethical considerations}
Recently, there has been a series of works attempting to utilize \ac{BCI} technology to enhance information accessing performance in various language-related applications, such as search~\cite{eugster2016natural,56pinkosova2020cortical,3allegretti2015relevance} and communication~\cite{pereira2018toward}. 
Such technology is currently at a very early stage where such applications feel a long way off.
However, it is important to discuss the associated concerns regarding privacy issues as the collection of brain signals is inherently susceptible to the actions of malicious third parties, which increases the risk of potential misuse or mishandling of sensitive information.

On the one hand, raw data collected via neurophysiological devices should be treated as private information, as such data can potentially be used to identify an individual~\cite{alsunaidi2020comparison} as well as their physiological disorders and thoughts~\cite{yin2022deep}. 
This technology may lead to risks such as influencing people's political opinions, and discrimination during recruiting based on their neural profiles. 
Therefore, the raw data should be avoided from being uploaded to the cloud for computation. 
It is necessary to filter sensitive information and decode only the information that helps the user accomplish their task with local computing. 
For publicly available datasets, ethical review and informed consent from each participant should be obtained, such as the dataset used in this paper~(see Section~\ref{sec:Dataset statistics}). 
Additionally, datasets should be used strictly for research purposes following their respective licenses.

On the other hand, there is a concern regarding the interaction log that might be recorded in applications like search engines.
Although such interactions, such as clicks, comments, and submitted queries, are frequently recorded for improving individual user experience, the utilization of BCI can potentially pose greater risks.
For example, it can be employed to capture users' genuine opinions on content within information systems, which can then be adopted in applications such as selective exposure and targeted advertising.
Hence, users should have the right to decide whether they are willing to provide their interaction history to service providers. 
This is already specified in the legislation of many countries.
In addition, the interaction history, even with users' permission, should undergo post-hoc filtering to remove any sensitive information before being utilized to train a model aimed at enhancing the commercial product.



\section{Reproducibility}
Our experiments use open-source data\-sets (Pereira's dataset~\cite{pereira2018toward}, Huth's dataset~\cite{lebel2023natural}, and the Narratives dataset~\cite{nastase2021narratives}, which can be downloaded from the paper websites or OpenNeuro\footnote{\href{https://openneuro.org/}{https://openneuro.org/}}). 
The data from \citet{pereira2018toward} is available under the CC BY 4.0 license.  
The Huth's dataset and Narratives dataset are provided with a ``CC0'' license. 
All code used in the paper are available under the MIT license~\footnote{\href{https://github.com/YeZiyi1998/Brain-Query-Augmentation}{https://github.com/YeZiyi1998/Brain-Query-Augmentation}}.

\clearpage
\bibliography{custom}

\clearpage
\appendix
\section{Appendix}
\label{sec:appendix}

\subsection{Dataset Information}
\label{sec:Dataset statistics}
Huth's dataset and the Narratives dataset both contain \ac{fMRI} responses recorded while participants listened to English auditory language stimuli of spoken stories.
Huth's dataset comprises data from 8 participants, with each participant listening to a total of 27 stories. 
As a result, each participant contributed approximately 6 hours of neural data, amounting to 9,244 \acp{TR}, i.e., the time frames for fMRI data acquisition.
On the other hand, the Narratives dataset initially included a total of 365 participants. 
However, due to the significantly high computational demand, we selected a subset of 8 individuals who had engaged in at least 4 stories, with an average of 2,109 \acp{TR} collected from each participant. 
Pereira's dataset collects participants' fMRI signals while viewing English visual stimuli composed of Wikipedia-style sentences.
In line with previous research by \citet{luo2022cogtaskonomy}, we selected cognitive data from participants who took part in both experiments 2 and 3. 
This subset consists of 5 participants, each of whom watched 627 sentences selected from 177 passages.
Each sentence corresponds to one \ac{TR}, which represents one scan of \ac{fMRI} data consisting of signals from approximately 10,000 to 100,000 voxels.
The statistics of these datasets are provided in Table~\ref{tab:data_static}.
All datasets received approval from ethics committees and are accessible for research purposes. 
We present the overall statistics of the above three fMRI datasets in Table~\ref{tab:data_static}.

\begin{table*}[t]
\centering
\small
\setlength{\tabcolsep}{1.6mm}
\begin{tabular}{cccccccc}
\toprule
\textbf{Dataset}    & \makecell[c]{\textbf{\#Partic-}\\\textbf{ipants}} & \makecell[c]{\textbf{\#Total} \\ \textbf{duration}} & \makecell[c]{\textbf{\#Duration per} \\ \textbf{participant}} & \makecell[c]{\textbf{\#Total} \\ \textbf{TRs}} & \makecell[c]{\textbf{\#TRs per} \\ \textbf{participant}} & \makecell[c]{\textbf{\#Total} \\\textbf{words}} & \makecell[c]{\textbf{\#Words per} \\\textbf{participant}} \\ 
\midrule
Pereira's   & 5   &7.0 h      &      1.4 h        &         3,135    &         627          &      38,650         &   7,730                  \\ 
\midrule
Huth's   & 8          & 3.5 days         & 10 h               &       122,992      &         15,374         &     427,296          &    53,412                 \\ 
\midrule
Narratives  & 8  &  7.5h    &     56 min         & 16,868      & 2,109      &    80,160     &  10,020   \\
\bottomrule
\end{tabular}
\caption{Overall statistics of \ac{fMRI} datasets.}
\label{tab:data_static}
\end{table*}

\begin{table*}[t]
\centering
\small
\begin{tabular}{cccccc}
\toprule
\textbf{Dataset}    & 
\multicolumn{1}{c}{\textbf{\#Query}} & \multicolumn{1}{c}{\textbf{\#Document}} & \textbf{Query length} & \multicolumn{1}{c}{\textbf{Continuation length}} & \multicolumn{1}{c}{\textbf{Doc length}} \\
\midrule
Pereira's    & 1,254    & 168        & 5.8$\pm$2.5      & 4.5$\pm$1.5             & 46$\pm$6       \\
Huth's       & 26,578   & 876        & 10.3$\pm$4.3     & 7.4$\pm$0.5             & 61.2$\pm$13    \\
Narratives & 4,979    & 162        & 9.5$\pm$4.7      & 6.0$\pm$1.9               & 60.0$\pm$23.5   \\
\bottomrule
\end{tabular}
\caption{Overall statistics of the document corpus and query set constructed with the fMRI datasets.}
\label{tab:data_static2}
\end{table*}

\subsection{Dataset preprocessing}
\label{sec:Dataset preprocessing}
\paragraph{Document corpus construction}
Pereira's dataset has a natural segmentation of documents, with approximately 3 to 4 sentences per document. 
Therefore, we utilized its inherent segmentation for our experiment.
After defining the document corpus, we utilize the same protocol to select a query in the ICT task and the next token prediction task construction.
So each query $Q$ is either a piece of sentence in Pereira's dataset or a text span corresponding to a \ac{TR}.
For Huth's dataset and the Narratives dataset, the language stimuli are presented continuously without any natural document segmentation provided.
Hence, we segment text spans presented in every 10 consecutive TRs as a document.
This segmentation criterion results in an average document length similar to the passage length found in existing IR benchmarks, such as MS MARCO~\cite{bajaj2016ms}~(see Section~\ref{sec:Dataset statistics} for detailed statistics). 
According to the segmentation, the average document length is about 60, which is similar to the passage length of existing IR datasets, like MS MARCO~\cite{bajaj2016ms}, which was used to train our baseline RepLLaMA.

\paragraph{Query construction}
Following existing research in language decoding from brain signals~\cite{tang2023semantic,ye2023language}, we split the text stimuli to construct the query according to the \ac{TR}.
For Pereira's dataset, we split each sentence into three parts with equal length. 
Two unique data samples are constructed by treating 
(i)~the first third as the query and the second third as the ground truth continuation as well as 
(ii)~combining the first two thirds as the query and using the last third as the ground truth continuation. 
For Huth's dataset and the Narratives dataset, we segmented the data by considering the perceived textual content during each TR as the ground truth continuation. 
We then truncated the preceding text and used it as the query.
The truncation is accomplished using a sliding window ranging from 1 to 3 TRs to pick the language stimuli.
We detail the average length of the queries, the query continuations, and the length of documents in Section~\ref{sec:Dataset statistics}.
The statistics of the query generation task and the document ranking task are presented in Table~\ref{tab:data_static2}.

\subsection{Query performance features}
\label{sec:Query performance features}
To study the effect of brain signals in query augmentation in queries with different features.
We analyze the document ranking performance according to the original queries measured by the following features:

(1)~Averaged ICTF~(inverse collection term frequency)~\cite{carmel2010estimating}:
ICTF is a popular measure for the relative importance of the query terms and is usually measured by the following formulas:
\begin{equation}
    ICTF(w) = log(\frac{\mid D \mid}{\mathit{TF}(w, D)} )
\end{equation}
where $\mid D \mid$ is the number of all terms in collection $D$, and $TF(w, D)$ is the term frequency~(number of occurrences) of term $w$ in $D$.
Here we use the averaged ICTF of all terms $w$ in the query.

(2)~Averaged IDF~(inverse document frequency)~\cite{hauff2008improved}:
IDF is another widely used measure for the importance of the query terms and is typically measured by the following formulas:
\begin{equation}
    IDF(w) = log(\frac{N}{N_w})
\end{equation}
where $N$ is the number of documents in the collection and $N_w$ is the number of documents containing the term $w$.
Here we use the averaged IDF of all terms $w$ in the query.


(3)~Specificity~(or simplified clarity score)~\cite{cronen2002predicting}:
Specificity score measures the Kullback-Leibler divergence of the query's language model from the collection's language model, which can be formulated as:
\begin{equation}
    \mathit{q} = \sum_{w \in q} P(w\mid q)log(\frac{P(w\mid q)}{P(w \mid D)})
\end{equation}
where $P(w\mid q)$ and $P(w \mid D)$ indicate the token possibility in the query and the document, respectively. 

(4)~Clarify~\cite{cronen2002predicting}:
Clarify score quantifies the ambiguity of a query w.r.t. a collection of documents.
It measures the KL divergence between a relevance model induced from top-ranked documents retrieved by the original query.
\begin{equation}
\small
    \mathit{Clarify}(q,D^k_{q:M})=\sum_{w \in V} P(w\mid D^k_{q:M})\frac{P(w\mid D^k_{q:M})}{P(w\mid D)}
\end{equation}
where $w$ and $V$ denote a query term and the entire collection vocabulary, respectively, $D^k_{q:M}$ indicates the top-k document retrieved by model $M$ using query $q$. 
The conjecture suggests that a larger KL divergence corresponds to a more clarified query and a better retrieval quality.

\begin{table*}[t]
    \centering
    \small
    \begin{tabular}{c l m{0.2\textwidth} m{0.4\textwidth} c}
    \toprule
    \multicolumn{1}{c}{\textbf{Dataset}} & \multicolumn{1}{c}{\textbf{Method}} & \multicolumn{1}{c}{\textbf{Query Content}} & \multicolumn{1}{c}{\textbf{Top-ranked document}} & \textbf{Relevance} \\
    \midrule
 \multirow{10}{*}{Huth's}   & \multicolumn{1}{c}{Original} & with \textcolor{true_blue}{one hand tied behind} & cup holder and gets ready to \textcolor{true_blue}{hand} him some change and ... if he got a cellphone I gotta get \textcolor{true_blue}{one} ...  & 0 \\
    \cmidrule{2-5}
   &  \multicolumn{1}{c}{Unsup-Aug} & with \textcolor{true_blue}{one hand tied behind} \textcolor{myPurple}{ my eyes shut} & ... like we're gonna hit and  \textcolor{myPurple}{I} just did the only thing \textcolor{myPurple}{I} thought seemed right \textcolor{myPurple}{I} just \textcolor{myPurple}{shut my eyes} ... & 0 \\
    \cmidrule{2-5}
   &  \multicolumn{1}{c}{RS Brain} & with \textcolor{true_blue}{one hand tied behind} \textcolor{myPurple}{thinking and what he's gonna} & ... \textcolor{myPurple}{he} just yells to me \textcolor{myPurple}{his} like we're \textcolor{myPurple}{gonna} hit and I just did the only thing I \textcolor{myPurple}{thought} seemed right I just shut my eyes I took a deep & 0 \\
    \cmidrule{2-5}
   & \multicolumn{1}{c}{Brain-Aug} & with \textcolor{true_blue}{one hand tied behind} \textcolor{myPurple}{my back and I'm thinking} & \textit{[MASK]} \textcolor{myPurple}{my back} which \textcolor{myPurple}{I} only probably ever would have to do with ... they were a \textcolor{true_blue}{handful} she was paying ten dollars an hour in nineteen eighty eight \textcolor{myPurple}{I} kind of \textcolor{myPurple}{thought} that all of \textcolor{myPurple}{my} & 1 \\
   \midrule
   \multirow{12}{*}{Narratives}   & \multicolumn{1}{c}{Original} & \textcolor{true_blue}{you get undressed and get into} &  gentlemen \textcolor{true_blue}{you} can't \textcolor{true_blue}{get} away with this sooner or later somebody the or somebody is going to \textcolor{true_blue}{get} wind of this madness ... & 0 \\ \cmidrule{2-5}
   & \multicolumn{1}{c}{Unsup-Aug} & \textcolor{true_blue}{you get undressed and get into} \textcolor{myPurple}{somebody going away} &  gentlemen \textcolor{true_blue}{you} can't \textcolor{true_blue}{get} \textcolor{myPurple}{away} with this sooner or later somebody the or \textcolor{myPurple}{somebody} is going to get wind of this madness ... & 0 \\ \cmidrule{2-5}
   & \multicolumn{1}{c}{RS Brain} &  \textcolor{true_blue}{you get undressed and get into} \textcolor{myPurple}{the bathtub and I'll wash} & \textcolor{true_blue}{you} just come with \textcolor{myPurple}{me} where \textcolor{true_blue}{into} the tunnel \textcolor{myPurple}{I'll} show you henry swanson led guy to a small hole on the ... & 0 \\ \cmidrule{2-5}
   & \multicolumn{1}{c}{Brain-Aug} & \textcolor{true_blue}{you get undressed and get into} \textcolor{myPurple}{ bed and I'll join you} & ... now Arthur listen \textcolor{myPurple}{I} say this in all sincerity will \textit{[MASK]} \textcolor{myPurple}{bed} like a good guy and relax ... & 1 \\
    \bottomrule
    \end{tabular}
\caption{Examples of document ranking with BM25 using the original query or the augmented query in Huth's and Narratives dataset. Text in \textcolor{true_blue}{blue} and in \textcolor{myPurple}{purple} indicates content in the original query and generated by the query augmentation method, respectively. \textit{[MASK]} indicates the position of the selected query in the ICT setting. }
\label{tab:case study appendix}    
\end{table*}

\subsection{Implementation Details}
\label{sec:implementation details}
To efficiently manage and analyze the high-dimensional fMRI data, we employ two methods to reduce dimensionality.
For Huth's dataset and Narratives dataset, we select features from brain regions identified by \citet{musso2003broca}, which are known to be relevant to language processing in the human brain.
For Pereira's dataset, we apply component analysis~\cite{abdi2010principal} on the original \ac{fMRI} features to reduce the dimensionality to 1000.
The 7B version of the Llama-2 model~\cite{touvron2023llama} released in Huggingface~\footnote{\href{https://huggingface.co/models}{https://huggingface.co/models}} is adopted as the language model for generating the query continuation.

We train Brain-Aug with the Adam optimizer~\cite{kingma2014adam} using a learning rate of $1\times 10^{-4}$ and a batch size of 8. 
The learning rate is selected from the set $\{1\times10^{-3},1\times10^{-4},1\times10^{-5}\}$ based on the experimental performance on Pereira's dataset.
The training of the warm-up step is stopped after ten epochs, while an early stop strategy was adopted in the training of the next token prediction task when no improvement was observed on the validation set for ten epochs. 
The entire training process was conducted on 16 A100 graphics processing units with 40 GB of memory and took approximately 12 hours to complete.
During the inference stage, we utilize a beam search protocol with a width of 5.

When performing query generation for document ranking, we set the maximum number of words that can be expanded to 5. 
In Pereira's dataset, the continuation will be 5 tokens unless the model generates a token indicating the end of the continuation. 
In the other two datasets, due to their higher perplexity, the model may generate content with lower quality. 
Therefore, during the generation process, we calculate the perplexity of the content generated up to the current step~(note that this is the perplexity of the generated content, not the ground truth label). 
If the averaged perplexity at the current step exceeds a threshold of 1.5, the generation process is early stopped.
Due to the fact that the queries constructed based on the above method can be quite long in Huth's dataset and the Narratives dataset, in order to simulate real-world query submission scenarios, we sampled 3 query terms from the original queries when evaluating the ranking performance.

\subsection{Example cases}
\label{sec:appendix_example}
We present the manually selected example cases in Huth's and Narratives's dataset in Table~\ref{tab:case study appendix}.
In these cases, Brain-Aug leverages brain signals and ranks the relevant document as top-1. 
The selection of these examples was based on the higher NDCG@1 scores of the Brain-Aug compared to the baselines and controls.
More cases can be found in the provided repository. 

\subsection{Failures and Insights}
In our research, we have also conducted two meaningful attempts, despite being unsuccessful, may provide insights for further research. 
The first attempt was to explore whether \ac{EEG} signals can be utilized for Brain-Aug, as EEG signals are easier to collect in real-world scenarios than \ac{fMRI}. 
However, we found that in our experiment with two public EEG datasets, i.e., UERCM~\footnote{\href{https://github.com/YeZiyi1998/UERCM}{https://github.com/YeZiyi1998/UERCM}} and Zuco~\footnote{\href{https://osf.io/2urht/}{https://osf.io/2urht/}}, Brain-Aug did not outperform RS Brain. 
This implies that the existing quality of EEG data have limitations in their ability to decode semantics with Brain-Aug.
The second attempt was to train a query augmentation model with brain signals to directly facilitate the document ranking task.
We constructed the unified prompts using the same method of Brain-Aug and fed them into Rep\-llama to obtain query representations. 
Then, we used a contrastive loss function to make these representations closer to the relevant documents.
We found that training the model in this way makes it challenging to generalize the performance to the validation set. 
This could be potentially attributed to the label-inefficient issue in dense retrieval training settings. 
Future research can further explore this direction.

\subsection{AI assistants usage}
After completing the paper, we employ ChatGPT\footnote{\href{https://chat.openai.com/}{https://chat.openai.com/}} and Gemini\footnote{\href{https://gemini.google.com/app}{https://gemini.google.com/app}} to identify writing typos. 
Subsequently, manual review and revision are performed to address these typos.

\end{document}